\documentclass{article}

\usepackage{authblk}
\usepackage{graphicx}
\usepackage{amsmath}
\usepackage{amssymb}
\usepackage{cite}
\usepackage{natbib}
\usepackage{url}
\usepackage{fullpage}

\usepackage{algpseudocode}
\usepackage[ruled, linesnumbered]{algorithm2e}
\usepackage{stmaryrd}

\SetCommentSty{commentfont}
\usepackage{amsmath}

\DeclareMathOperator*{\argmax}{arg\,max}
\usepackage{multicol}
\usepackage{multirow}

\usepackage{booktabs} 

\newcommand{\mname}{\texttt{HAMLET}\xspace}

\begin{document}
\title{\mname: Interpretable Human And Machine co-LEarning Technique}
\date{March 26, 2018}

\author[1]{Olivier Deiss}
\author[1]{Siddharth Biswal}
\author[2]{Jing Jin}
\author[2]{Haoqi Sun}
\author[2]{M. Brandon Westover}
\author[1]{Jimeng Sun}
\affil[1]{Georgia Institute of Technology}
\affil[2]{Massachusetts General Hospital}

\renewcommand\Authands{ and }


\maketitle





\begin{abstract}
Efficient label acquisition processes are key to obtaining robust classifiers. However, data labeling is often challenging and subject to high levels of label noise. This can arise even when classification targets are well defined, if instances to be labeled are more difficult than the prototypes used to define the class, leading to disagreements among the expert community.

Here, we enable efficient training of deep neural networks. From low-confidence labels, we iteratively improve their quality by simultaneous learning of machines and experts. We call it Human And Machine co-LEarning Technique (\mname). Throughout the process, experts become more consistent, while the algorithm provides them with explainable feedback for confirmation. \mname uses a neural embedding function and a memory module filled with diverse reference embeddings from different classes. Its output includes classification labels and highly relevant reference embeddings as explanation. 


We took the study of brain monitoring at intensive care unit (ICU) as an application of \mname on continuous electroencephalography (cEEG) data. 
Although cEEG monitoring yields large volumes of data, labeling costs and difficulty make it hard to build a classifier. Additionally, while experts agree on the labels of clear-cut examples of cEEG patterns, labeling many real-world cEEG data can be extremely challenging. Thus, a large minority of sequences might be mislabeled.
\mname has shown significant performance gain against deep learning and other baselines, increasing accuracy from 7.03\% to 68.75\% on challenging inputs. Besides improved performance, clinical experts confirmed the interpretability of those reference embeddings in helping explaining the classification results by \mname.

\end{abstract}






\maketitle





\section{Introduction}

For a wide spectrum of real-world applications, ranging from image classification (K. He et al., 2015~\cite{DBLP:journals/corr/HeZRS15}), face and speech recognition (A. Y. Hannun et al., 2014~\cite{DBLP:journals/corr/HannunCCCDEPSSCN14}), and bioinformatics (A. Esteva et al., 2017~\cite{Esteva2017}), to speech synthesis (A. van den Oord et al., 2016~\cite{DBLP:journals/corr/OordDZSVGKSK16}) and even game playing (V. Mnih et al., 2015~\cite{Mnih2015}), deep learning has become one of the most powerful tools. Convolutional Neural Networks (CNNs), which are biologically-inspired models (Y. LeCun et al., 1989~\cite{6795724}) able of reaching great performance in image classification (A. Krizhevsky et al., 2012~\cite{Krizhevsky:2012:ICD:2999134.2999257}), natural language processing (W. Yin et al., 2017~\cite{DBLP:journals/corr/0001KYS17}) or analysis of time-series data, are a great example of successful models. However in supervised learning, excellent performance always comes with the same burden, regardless of the field of research: the need of large quantities of high-quality labeled data. The problem of detecting events in electroencephalograms (EEG) belongs to this class of applications where deep learning can achieve excellent results but requires large amounts of labeled data.  


However, in many real world applications, the labels are difficult to acquire. 
Either the acquisition costs are too high to make it possible to collect enough data, or events of a certain label are simply too rare to be observed enough times. In both situations, it becomes hard to apply deep learning algorithms. In other recurring situations, there is an abundance of raw data, but a lack of high quality labeled data, again due to either high labeling costs or the difficulty of the labeling task. In biomedical applications, data acquisition is a first challenge to be overcome.
First, for privacy reasons, it can be difficult to obtain patient data. Second, labeling is often expensive in that it requires availability of a domain expert who can dedicate enough time to the dataset creation. Third, tasks can become challenging to the point where even domain experts cannot readily come to an agreement on some or many sample points (N. Gaspard et al., 2014~\cite{gaspard2014interrater}). For a classification task, they might often disagree on difficult sample points at the boundaries of multiple classes, or have different interpretations of the established concepts. For these reasons, obtaining a large dataset with high-quality labels  is quite challenging.

As an example, one great challenge in biomedicine is related to the classification of seizures, which are the result of abnormal electrical activity in the brain. There exist multiple types of seizures, and classifying them allows to study their respective impact on health, as well as how to effectively cure or treat them. A recent study by Ruiz et al.~\cite{doi:10.1001/jamaneurol.2016.4990} showed that the analysis of continuous electroencephalography (cEEG) signals can help predict the risk of seizures in critically ill patients. cEEG is a non-invasive method to monitor the electrical activity of  the brain. In the critical care setting, cEEG monitoring is typically performed for 24-72 hours at a time, providing large volumes of data. However, manually labeling the events is a tedious task for the human expert, both due to the high volumes of data and the difficulty of the task.

Deep learning models could advance both the clinical value of brain monitoring and provide valuable scientific tools for studying seizures and related pathological cEEG events. Deep neural networks have already been used to study EEG patterns (P. W. Mirowski et al., 2008~\cite{4685487}; P. Bashivan et al., 2015~\cite{DBLP:journals/corr/BashivanRYC15}), so they are a model of choice in our study. The issue here is the costs related to labeling data, making it hard to rely on a standard learning framework. In addition, EEGs are difficult for experts to label consistently due to the frequent overlap between classes (JJ. Halford et al., 2015~\cite{halford2015inter}; H. A. Haider et al., 2016~\cite{haider2016sensitivity}). The standard solution to this problem in the medical literature is to have each sample reviewed and labeled independently -- or in a committee -- by multiple human experts before a decision on the class label is taken. This approach is however not  scalable. Overall, this situation makes it challenging to obtain a labeled dataset suitable for proper training of deep neural networks, and therefore is a great application of our work on label acquisition. 

With the help of active learning, it has been shown that a classifier can be trained on a judiciously selected subset of the available data while performing as accurately as models trained with a much larger set of randomly selected training examples (K. Wang et al., 2017~\cite{DBLP:journals/corr/WangZLZL17}). Such efficient methods for selecting sample points for human labeling allow to cope with budget restrictions with no or limited trade-off on performance. However, although a similar framework could help us efficiently grow our dataset, all active learning methods start with the assumption that they can query an ``Oracle'', a human expert who can label any sample point -- the query -- with no risk of misclassification. In many studies, the ``Oracle'' is a human, expert in the field, and the correct class label leaves no room for doubt. In other studies, like ours, the problem is so challenging that fully relying on human experts in order to obtain ground truth class labels is not enough. Indeed, the task is so difficult that there is a great risk that the label given by the human expert may still be wrong. Therefore, although active learning remains attractive for our study, using it would not solve our issues.

Instead, we first acknowledge that in an increasing number of studies, although the model can be seen as being taught by humans, it is not uncommon that the overall accuracy of the model is better than that of all human teachers combined (V. Gulshan et al., 2016~\cite{doi:10.1001/jama.2016.17216}; D. Silver et al., 2017~\cite{Silver2017}). In this work, we use this fact as a way to improve the human raters' performance and consistency in the labeling task, especially when labeling data that spans multiple different patients, and we propose \mname, a Human And Machine co-LEarning Technique to efficiently bypass this label acquisition difficulty. We apply \mname to train a classifier for various types of non-convulsive seizures, based on continuous EEG recordings. \mname helps us face the above mentioned issues, while making use of the great performance of deep learning models at challenging tasks to improve label quality in a feedback-loop fashion. \mname fully integrates the limiting constraint of lacking an ``Oracle'' or absolute source of truth. With \mname, we were able to improve our dataset, ultimately obtaining higher classification accuracies. Our contributions can be summarized as follow:
\begin{itemize}
  \item Design of an algorithm for efficient label improvements.
  \item Increased interpretability of our classifier with the use of a separate memory module hosting representative reference embeddings.
  \item Successful application of our method to the challenging task of seizure classification.
\end{itemize}

In this article, we first survey related work in EEG classification, deep learning for health informatics, active learning and co-training. Our co-learning technique and the architecture of our classifier are introduced in section~\ref{sec:model}. Finally, in section~\ref{sec:experiments}, we present our dataset, methods for pre-processing, and results.




\section{Related Work}
\label{sec:related}

  \subsection{EEG Classification}

    \subsubsection{Using Feature Extraction}

The topic of EEG classification is a fertile area of research. Most methods traditionally rely on feature selection combined with a classifier, such as Support Vector Machines (SVM), Linear Discriminant Analysis (LDA) or $k$-Nearest Neighbors ($k$-NN) (M. H. Alomari et al., 2013~\cite{DBLP:journals/corr/AlomariSA13}; H. Shoeb et al., 2010~\cite{aliseizure}).
Relevant features include Common Spatial Patterns (CSP), Filter-Bank Common Spatial Patterns (FBCSP), and Logarithmic Band Power (BP) (X. Yong et al., 2015~\cite{10.1371/journal.pone.0121896}). Seizure classification has also been approached with feature engineering (F. F{\"u}rbass et al., 2015~\cite{furbass2015automatic}) for instance using signal amplitude variation (AV) and a regularity statistic (PMRS) (J. C. Sackellares et al., 2011~\cite{sackellares2011quantitative}).
Although careful feature engineering can lead to great performance, it is not be strictly necessary.

    \subsubsection{Deep Learning}

Classification of EEG data can be seen as a multivariate time-series classification problem (P. Bashivan et al., 2015~\cite{DBLP:journals/corr/BashivanRYC15}). 
Furthermore, one  advantage of CNNs is the automated feature selection that happens during the training process. Without additional work, the model learns the features that it finds most relevant for its given task, from the raw signals given as input. This has been shown with great success for sleep staging (S. Biswal et al., 2017~\cite{DBLP:journals/corr/BiswalKSGWBS17}) as well as seizure classification (U. Rajendra Acharya et al., 2017~\cite{ACHARYA2017}). However, all these deep models require high volumes of labeled data for training, which is the bottleneck in our application. Therefore, CNNs remain a model of choice but we cannot use them directly in the present study due to the label limitation.

  \subsection{Convolutional Auto-Encoders (CAEs)}

CNNs can be used in a supervised learning framework, but cannot benefit from the large amounts of unlabeled data. Traditionally, auto-encoders (AEs) have been widely used to extract meaningful features in the absence of labels even on biomedical data (J. Tan et al., 2015~\cite{breastcancer}). However, in their simplest form, AEs cannot be efficiently applied to time-series as they ignore their bi-dimensional structure. Convolutional Auto-Encoders (CAEs) bring together the advantages of using convolutions and auto-encoders (X. Mao et al., 2016~\cite{NIPS2016_6172}; Masci et al.~\cite{Masci2011}). However, seizures are typically relative rare events, whereas CAEs tend to learn to represent the dominant statistical structure of the underlying data, which is not directly relevant for seizure classification. Therefore using CAEs alone is not sufficient in our classification task.

  \subsection{Co-Training}

Co-training is a semi-supervised machine learning algorithm, where two models can be concurrently trained, each with its own set of features. Additionally, the predictions of one model on an unlabeled dataset are used to enlarge the dataset of the other model. The algorithm has originally been introduced for the classification of web pages (A. Blum et al., 1998~\cite{Blum:1998:CLU:279943.279962}). Here we do not focus on increasing dataset size but rather label quality within the labeled dataset. Although the standard setting for co-training includes unlabeled data that models can benefit from, it has been shown that co-training can still save labeled samples even in a strictly supervised setting (M. Darnst{\"a}dt et al., 2009~\cite{10.1007/978-3-642-24412-4_33}) which we use. In the original co-training algorithm, there is the possibility that the training set of a model is augmented with wrong labels from the other model. COTRADE improves on this idea by only allowing models to share labels they are confident about (M. L. Zhang et al., 2011~\cite{5910412}). This notion of labeling confidence is at the center of \mname, however, there are some differences in our approach to co-training. Indeed, one of the two entities at the center of the algorithm is the human expert. The expert knowledge helps training the algorithm by updating the labeled dataset, while the model feedback helps improving the consistency of the expert at the classification task. Because of the human learning dimension, we use the term co-learning instead.

  \subsection{Active Learning}

In situations where unlabeled data is abundant, but labeled data is scarce due to a restricted labeling budget, active learning offers a way to select the most informative sample points in order optimize labeling resources on the data that is most helpful for the model. Heuristics for active learning include Query by Committee (QBC), uncertainty sampling, margin sampling, entropy or Expected Gradient Length (EGL) (B. Settles, 2010~\cite{Settles10activelearning}). Although these methods might reduce the need for large label sets, this still is not enough for deep learning models like CNNs. For such models, including pseudo-labeled high-confidence data into the training set is a possible approach (K. Wang et al., 2017~\cite{DBLP:journals/corr/WangZLZL17}). Unfortunately, active learning implies the existence of a source of truth that is not available in our challenging classification task, since in our study, even humans find it difficult to label many of the patterns commonly encountered in EEGs of critically ill patients.

  \subsection{Using Memory Modules in Neural Networks}

Recently, researchers have been experimenting with augmenting neural networks with memory modules. For instance, a memory module is present in the architecture of Matching Networks (O. Vinyals et al., 2016~\cite{DBLP:journals/corr/VinyalsBLKW16}). In their model, the module is used together with an attention mechanism, however without interpretability in mind. Our design, presented in the following section, emphasizes interpretability thanks to an external memory module.




\begin{figure}
\center
\includegraphics[width=3in]{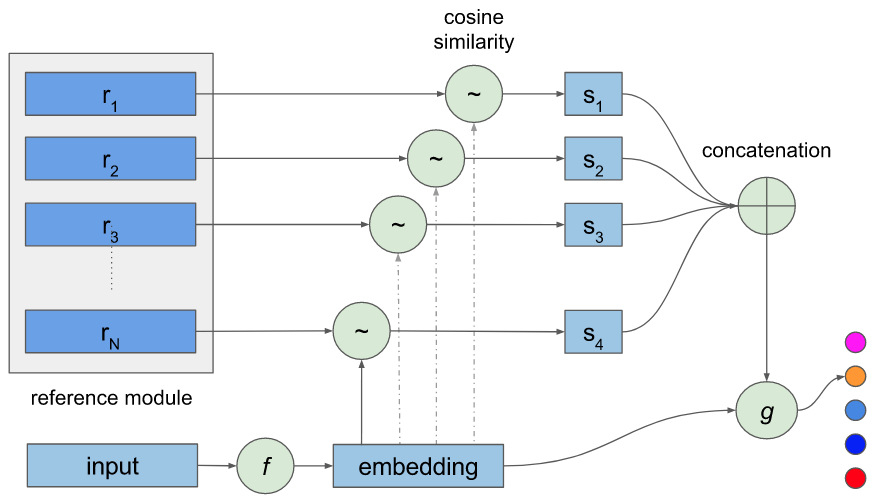}
\caption{Model Structure. $f$ is an embedding function, either a CNN (supervised training) or a CAE (unsupervised). The memory module pictured above the input contains reference embeddings used to compute similarity scores $s_i$. $g$ is the final classification layer, typically a dense layer.}
\label{fig:model}
\end{figure}

\section{\mname Method}
\label{sec:model}

  \subsection{Co-learning Framework}
  
    \subsubsection{Architecture of the Classifier}
  
One important part of our \mname framework is the classifier, that is used to both improve the human expert through re-labeling suggestions, and of course the classification task itself, the ultimate goal of the whole process. Its architecture is shown in figure~\ref{fig:model}. It is based on a combination of the following three components:
\begin{itemize}
  \item Embedding function $f$
  \item Memory module containing $N$ reference embeddings
  \item Dense layer $g$
\end{itemize}

One key novelty of our model lies in what we call a memory module. In this separate module, a set of $N$ reference embeddings is stored. For a given input $i$, the encoder creates an embedding $e_i = f(i)$ that is compared to each reference embedding $r_j$ using cosine similarity, giving $N$ similarity scores $s_j$. These similarity scores and the current embedding $e_i$ are concatenated to form the immediate representation fed to the classifier function $g$, typically a multilayer perceptron (MLP). We will explain in further sections how using reference embeddings enhances interpretability.

In our experiments, we used $N = 512$, and a dense layer $g$ with one hidden layer of $n = 1024$ neurons. We propose two different embedding functions $f$: 1) a supervised alternative that is trained with the available labeled dataset; and 2) a Convolutional Auto-Encoder (CAE), which allows to benefit from the whole dataset.

    \subsubsection{Algorithm}

\begin{figure}
\center
\includegraphics[width=2.6in]{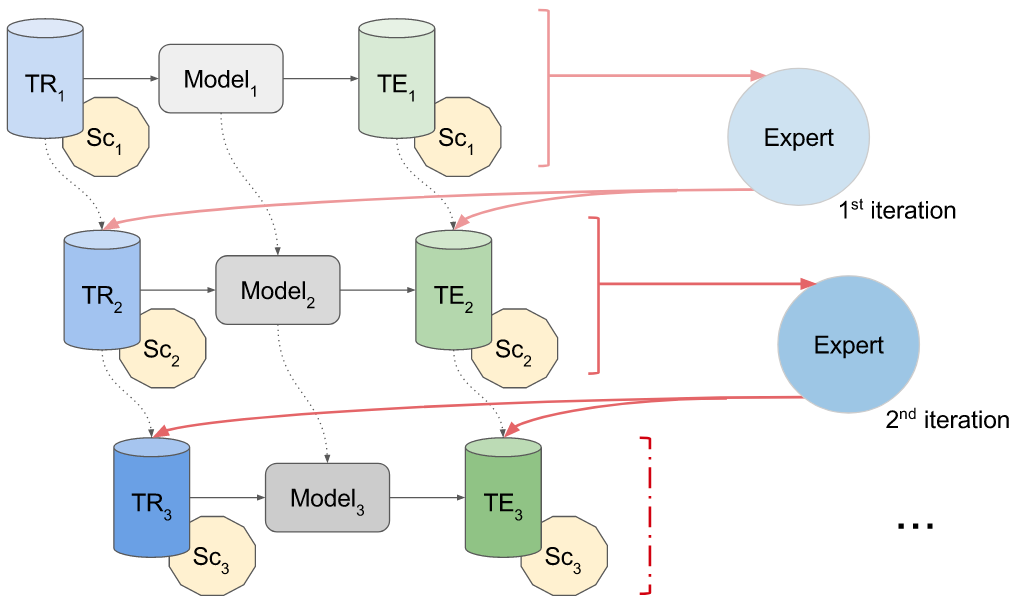}
\caption{Co-Learning algorithm with machine feedback. The model is first trained on the original training set TR$_1$ and evaluated on the testing set TE$_1$, keeping track of training and testing scores Sc$_1$. The expert evaluates the results, and updates a subset of both TR$_1$ and TE$_1$. A more robust model is obtained by fine-tuning the original one using the newer version of the dataset. Throughout the process, the model improves and the expert becomes more consistent.}
\label{fig:algo}
\end{figure}

Our co-learning framework is best described with the following algorithm. We start with a first dataset with low quality labels, and iterate through the following steps:
\begin{enumerate}
  \item Fine-tuning (pre-training for the first iteration) of the embedding function $f$.
  \item Selection of reference embeddings.
  \item Learning of the dense layer $g$.
  \item Label improvement through machine feedback.
\end{enumerate}
The procedure is illustrated on figure~\ref{fig:algo}. By going through multiple runs of the above algorithm, the model is iteratively improved thanks to the increase in label quality. Concurrently, the expert learns from past classification mistakes and becomes more consistent at the labeling task. Each phase of the algorithm is described in the following sections.

  \subsection{Fine-Tuning (or Pre-Training) of the Embedding Function $f$}
  \label{sub:modelstructure}

In this step, either a supervised or unsupervised approach can be taken. During the first iteration, the function $f$ is trained from scratch on the dataset. Further iterations of this algorithm only fine-tune $f$ and no pre-training is required.

    \subsubsection{Supervised Embedding Function $f$}
    \label{sec:cnn}

For the supervised approach, we experimented with a CNN. In this case, $f$ corresponds to the first layers of the model up until the start of the dense layer. After training is complete, the dense layer is dropped.
  
The architecture of our CNN is shown at the top of figure~\ref{fig:arch}. The first layer uses a depth-wise separate convolution in order to better handle the high dimensionality of input data in EEG studies -- 16 montages -- a practice inspired by recent work on EEG classification (R.T. Schirrmeister et al., 2017~\cite{DBLP:journals/corr/SchirrmeisterSF17}). The depth-wise separate convolution first performs a convolution through time (per channel), followed by a convolution across all electrodes (also known as \texttt{1x1} convolution). The following blocks are standard groups of convolution, max-pooling, and dropout layers. We have used a dropout rate of $p = 0.2$ and batch normalization, which has been shown to improve generalization (S. Ioffe et al., 2015~\cite{DBLP:journals/corr/IoffeS15}) right after the convolution layers during training. We use Exponential Linear Units (ELU) which provide faster learning (D-A Clevert et al., 2015~\cite{DBLP:journals/corr/ClevertUH15}) as our activation functions. The dense layer has 1024 neurons in the hidden layer, and ends with 5 softmax units corresponding to our class labels. We also use this CNN as a baseline for classification accuracy.

    \subsubsection{Unsupervised Embedding Function $f$}
    \label{sec:cae}

In the unsupervised approach, a Convolutional Auto-Encoder (CAE) is a great choice for $f$. A CAE is made of an encoder, which creates an embedding, and a decoder that takes this embedding and reconstructs the original sequence. After the CAE is trained on the whole dataset, the decoder is dropped, and $f$ is the encoder.
  
\begin{figure*}
\center
\includegraphics[height=8cm]{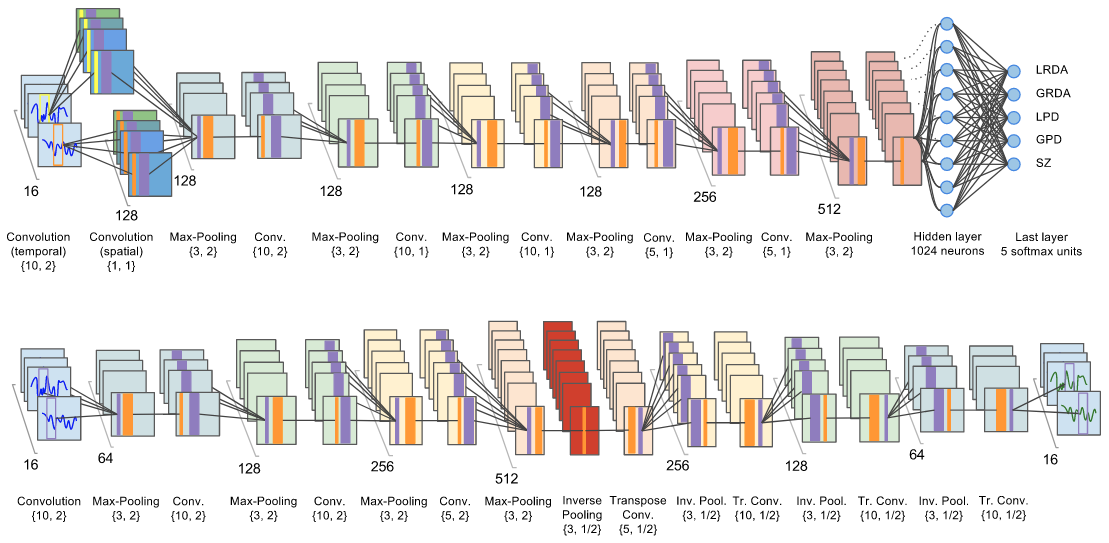}
\caption{Embedding functions $f$. Each input channel is 1D and parameters are indicated as \texttt{\{width, stride\}}. Top: supervised embedding function. Not shown on the figure, batch-normalization is applied after convolutions and dropout after max-pooling layers with a rate of $p = 0.2$. The dense layer is only used for training, then dropped when used in \mname. Bottom: unsupervised embedding function. The first layers (before red boxes) are the encoding part of the network. They build the embedding (shown in red). The following layers are part of the decoder, which is only used for training, then dropped when used in \mname.}
\label{fig:arch}
\end{figure*}

Our CAE architecture is shown at the bottom of figure~\ref{fig:arch}. In the CAE, there is no dense layer. Instead all the operations are reversed in the decoder after the last max-pooling layer. To build the decoder, we use un-pooling layers which increase the sequence size from encoded representation to input length, and transposed convolutions (sometimes called de-convolutions) in order to keep the overall structure completely symmetric.

  \subsection{Selection of the Reference Embeddings} 

The reference embeddings are selected right after the embedding function $f$ is learned. It is crucial that they are a close representation of the input data, and diverse enough to allow the model to use the module for all the classes present in the dataset. To ensure diversity, we use $k$-means among inputs of a same class, giving equal space in the memory module to all classes. Compared with a $k$-means on all available inputs, some reference embeddings might not be as useful for the model, or might even be redundant for rare classes, but this ensures that embeddings are always available for the model to learn from. For interpretability purposes, we keep track of the labels and sequences that correspond to the selected embeddings.

  \subsection{Learning of the Dense Layer $g$}

The next step is to teach the model to correctly classify sequences of EEG, using both the embedding function $f$ and the memory module. The outputs of both components are concatenated and fed to the classifier $g$. The model is then trained on the labeled dataset in a supervised manner.

  \subsection{Label Improvement with Machine Feedback}

The final step of training is where machine feedback comes into play. Using one of the machine feedback strategies described in~\ref{sec:feedbstrategies}, we suggest new labels for some sequences that the model mis-classified with high certainty. These suggestions are likely to propose sequences that have been mis-classified by the human grader. They are given to an expert for re-evaluation.

After a portion of the dataset has been re-labeled, the model is fine-tuned and new tests can evaluate performance increases. The labeling effort should be shared between samples from the training and testing sets. Although updating the testing set does not improve the model, it improves the quality of the experiment as a whole. A reasonable choice is to share this effort proportionally to each dataset size. However, in our experiments we also include results showing improvements on re-evaluated testing sequences only, to better display the improvements brought to the labels.

  \subsection{Machine Feedback Strategies}
  \label{sec:feedbstrategies}

In traditional active learning, various heuristics can be used for suggesting new data to be labeled, with the intent of increasing the labeled dataset size. This is called uncertainty sampling. These heuristics include Least Confidence (D.D. Lewis et al., 1994~\cite{Lewis:1994:SAT:188490.188495}), Margin Sampling (T. Scheffer et al., 2001~\cite{Scheffer:2001:AHM:647967.741626}) and entropy (C. E. Shannon, 1948~\cite{6773024}). Each heuristic outputs samples that would be most informative for the model. Here, we are looking for misclassification points that our model is most confident about, in order to suggest potential label errors by humans. Therefore we modify the active learning heuristics for improving label quality during the co-learning process. Below is a description of the various strategies for co-learning. 

    \subsubsection{Highest Confidence}

Using the confidence of the model accuracy on each piece of input data, we can select inputs that the model is most confident about. For each input $i$, the confidence $c_i$ is directly given by the probability of the class with highest probability according to the model: $c_{i} = \max_{j}{p(y_i = j)}$ for each class $j$. After all confidence values $c_i$ are obtained, they are ranked in decreasing order. For inputs $i$ with high confidence values $c_i$ that were wrongly classified by the model, a mistake on the original label is very likely. Therefore, those points will be provided as machine feedback to human experts for relabeling.

  \subsubsection{Margin Sampling}

The margin sampling heuristic is also based on confidence values. Inputs with the highest difference between the confidence values of their two most likely classes indicate high confidence. For input $i$: $m_{i} = p(y_i = j_1) - p(y_i = j_2)$, where $j_1$ and $j_2$ are the two most likely labels, according to the model. The misclassification points with large margin will be used as machine feedback.

  \subsubsection{Entropy}

Finally, entropy can also be used as a measure of the certainty for each input sequence. The entropy of all sequences are sorted in increasing order, the lowest values being the least informative ones -- i.e. those the model is most confident about. For a given input sequence $i$, the entropy $e_i$ is given by:
\begin{displaymath}
  e_{i} = -\sum_{j = 1}^{C}{p(y_i = j)\log(p(y_i = j)}
\end{displaymath}

  \subsection{Interpretability of the Model}

When using the model to perform a classification task on new data, the memory module can be used to explain the reasoning of the model. For a given input $i$, $N$ similarity scores $s_k$ will be generated from the memory module, one per reference embedding $r_k$.

In our experiments, we show that \texttt{\mname-CNN} learns to effectively use the reference embeddings for classification. However, we expect \texttt{\mname-CAE} to lack interpretability, as unsupervised training only teaches the model to recognize features that are only relevant to sequence encoding. As a result, \mname uses the memory module as a bank of features more than as a way to discriminate between classes. Therefore, interpretability can only be claimed for the supervised \texttt{\mname-CNN}.

Model interpretability allows to better justify the label suggestions in the last phase of the algorithm. For a given input sequence, if we look at the closest embeddings within the memory module -- those with highest similarity scores -- we will most likely reach a reference sequence that shows a similar pattern to the current one. Formally, to justify the decision of the model to classify $i$ into class $c$, we can look at the reference embedding $r*$ with highest similarity score among reference embeddings for the same class $c$:
\begin{displaymath}
r* = \argmax_{r_k\in c}s_k, \hspace{0.2cm} s_k = \text{cosine\_similarity}(r_k, f(i)).
\end{displaymath}

At this point, we can show the labels previously assigned to that reference sequence -- preferably by the same expert -- and explain why the model made such a suggestion. This increased interpretability enhances co-learning in the two following ways:
\begin{itemize}
  \item First, the expert knows the model did not randomly happen to output a given class label. This decision is supported by interpretability and is far less likely to be ignored by the expert.
  \item Second, by reminding experts how they previously labeled similar sequences, they can learn much faster and become more consistent while labeling. 
\end{itemize}

\section{Experiments}
\label{sec:experiments}


\begin{figure}
\center
\includegraphics[width=3.35in]{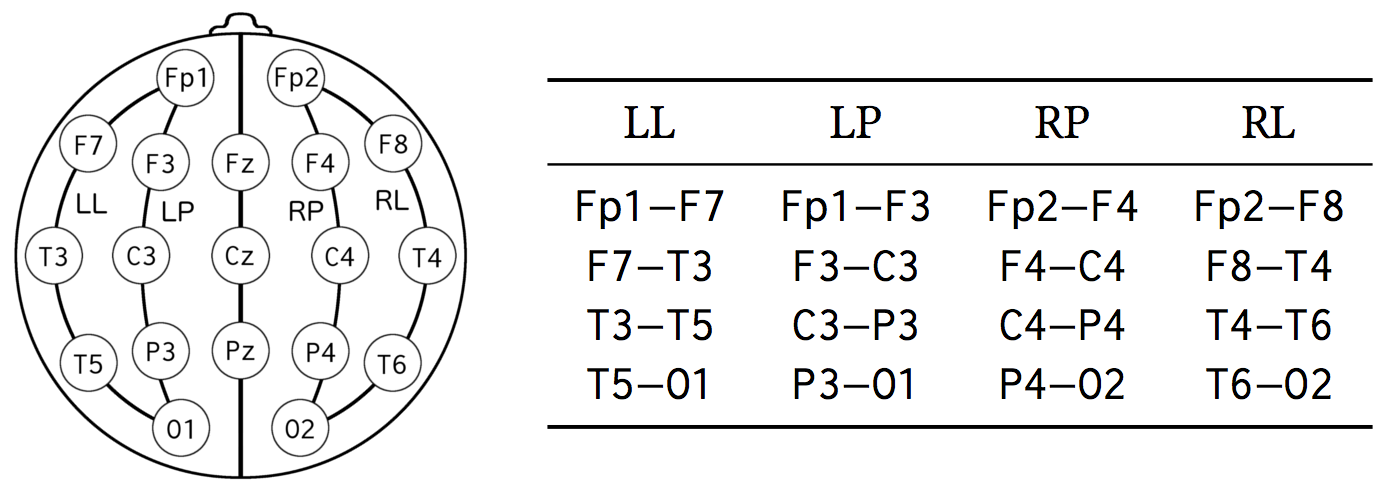}
\caption{Left: 19-channel EEG cap  showing the location of electrodes on the scalp, seen from above, as well as the different brain areas. \texttt{Fz}, \texttt{Cz} and \texttt{Pz} are reference electrodes. Right: bipolar montages. There are four montages for each of the brain areas: Left Lateral (LL), Left Posterior (LP), Right Posterior (RP), Right Lateral (RL). Each montage is the difference between the channels.}
  \vspace{0.5cm}
\label{fig:19channels}
\end{figure}

  \subsection{Dataset}
  \label{sec:dataset}

    \subsubsection{Acquisition of Continuous EEG Recordings}

EEG recordings are multivariate time-series describing the electrical activity of the brain. In this study, they are recorded in a non-invasive manner by affixing 19 small metallic (silver/silver chloride) electrodes directly to the scalp. The  EEG electrodes were placed in standardized locations, following the international 10-20 system, with locations and labels of electrodes as shown in figure~\ref{fig:19channels}.

Our original dataset, provided by the Neurosciences ICU at Massachusetts General Hospital (MGH), contains multiple hour-long (generally >24 hours) recordings of EEG for 155 different patients, sampled at $f = 200$ Hz. There is a plan to release the de-identified dataset to the public in the near future. In this dataset, we denote the EEG of patient $i$ by $X_i\in\mathbb{R}^{d_e\times m_i}$, with $d_e = 19$ the number of electrodes and $m_i$ the length of the given EEG.

    \subsubsection{Pre-processing}

After acquisition, the raw EEG data goes through the following pre-processing pipeline:
\begin{itemize}
  \item{Low-pass filtering:} the raw signals are first filtered with a 60.0 Hz low-pass filter, so that high-frequency noise and electrical artifacts are reduced.
  \item{Computation of montages:} it is usual to work with a bipolar montages instead of the raw data from the electrodes, to reduce interference from electrocardiogram (EKG) signals. This is done according to the table in figure~\ref{fig:19channels}. Let $M_i\in\mathbb{R}^{d_m\times m_i}$ be the montages for $X_i$ for patient $i$, with $d_m = 16$ the number of montages generated in the process -- i.e. channels.
  \item{Splitting of recordings:} all EEG recordings are split into 16-second sequences. We denote each sequence from $X_i$ and $M_i$ by $s_{i,j}$ and $m_{i,j}$, respectively. Because the patterns representative of each class are usually clear only within a larger contextual time window, experts look at an additional 6 seconds of signal on each side of the sequence when labeling.
\end{itemize}

    \subsubsection{Initial Labeling Process}

We have obtained the initial labeled recordings of 155 patients, for a total of 4176 hours, or an average of 27 hours per patient. Each sequence from these recordings has been manually labeled by a clinical expert. As we described about this task, high error rate is expected in these initial labels.

Each sequence of EEG can be given one of six class labels, where five correspond to the patterns of brain activity of primary interest (Seizure, Lateralized Periodic Discharges (LPD), Generalized Periodic Discharges (GPD), Generalized Rhythmic Delta Activity (GRDA), Lateralized Rhythmic Delta Activity (LRDA)), and the last one corresponds to Other/Artifacts (O/A). A great majority of the recordings is made of either background activity, noise, and non-physiological artifacts (O/A), so we put such sequences aside. In real-life, we can easily automate this selection step with a binary classifier. Let $D$ be the labeled dataset we obtain at this stage, without O/A ($D$ contains 5 classes).

\begin{table}
\center
  \caption{Number of 16-second sequences from each class in the full labeled dataset $D$.}
    \vspace{0.5cm}
  \label{table:histo}
  \begin{tabular}{lcc}
    \toprule
    Class Label&Sequence Count&Percentage\\
    \midrule
    Seizure&128,691&33\%\\
    LPD&115,729&30\%\\
    GPD&84,168&22\%\\
    GRDA&32,566&8\%\\
    LRDA&29,332&7\%\\
    \midrule
    Total&390,486&100\%\\
    \bottomrule
  \end{tabular}
\end{table}

    \subsubsection{Creation of Balanced Datasets}
    \label{sub:reduction}


For our experiments, we created three datasets from all the available 16-second sequences $s_{i,j}$ in $D$. The class distribution in the full dataset $D$ is highly skewed, as shown in table~\ref{table:histo}, so we keep the datasets balanced in terms of class labels to ensure the models do not learn trivial frequency bias:
\begin{itemize}
  \item{$D_{20k}^{unseen}$:} 20,000 sequences (89 hours of EEG data), split into a training set (80\%) and a testing set (20\%). Patients in the testing set are not present in the training set (testing is performed on \textit{unseen} patients).
  \item{$D_{20k}^{known}$:} 20,000 sequences (89 hours of EEG data), split into a training set (80\%) and a testing set (20\%). Patients in the testing set are also present in the training set (testing is performed on \textit{known} patients).
  \item{$D_{100k}$:} made of 100,000 sequences from $D$ that are not in the previous two datasets. This larger dataset represents 445 hours of recordings and is only used for unsupervised training of the embedding function $f$ (CAE). Patients in the testing set are not present in the training set.
\end{itemize}

    \subsubsection{Dataset Augmentation}

In this classification task, the most challenging issue is to make the model learn how to generalize across new patients. During training, in order to simulate different patients, the electrodes from the left and right side of the brain can be flipped, while the three reference electrodes in the middle of the scalp -- \texttt{Fz}, \texttt{Cz} and \texttt{Pz} -- remain unchanged. Our classification task being a symmetric problem, which particular side of the brain exhibits a pattern does not affect classification. This simple technique almost duplicates the training dataset in terms of the number of patients.

  \subsection{Setup}

The model has been trained using a server with Intel(R) Xeon(R) CPUs E5-2630 v3 running at 2.40 GHz, 32 cores, with 256 Gb of RAM and 4 GPUs Tesla K80, NVIDIA Corporation GK210GL, with CUDA v8.0. Training has been performed with Python version 2.7, using version 1.4.1 of \texttt{Tensorflow}. Other libraries needed for \mname include \texttt{numpy} and \texttt{scipy}. With the above configuration, training \texttt{\mname-CNN} during 100 epochs on $D_{20k}^{unseen}$, with a batch size of 128, took thirteen hours. We used the \texttt{Tensorflow} implementation of the stochastic optimizer \textit{Adam} (D.P. Kingma et al., 2014~\cite{DBLP:journals/corr/KingmaB14})

  \subsection{Evaluation}
  \label{sec:evaluation}

\begin{table}
\center
  \caption{Accuracy obtained on testing set of $D_{20k}^{known}$ (known patients) or $D_{20k}^{unseen}$ (unseen patients). Classifying EEGs of new patients is significantly more challenging.}
    \vspace{0.5cm}
  \label{tab:accuracy}
  \begin{tabular}{ccc}
    \toprule
    Model&Unseen Patients&Known Patients\\
    \midrule
    \texttt{\mname-CNN}&39.36\%&75.91\%\\
    \texttt{\mname-CAE}&38.46\%&75.07\%\\
    \texttt{CNN}&38.89\%&74.93\%\\
    \texttt{MLP}&21.04\%&20.05\%\\
    \bottomrule
  \end{tabular}
\end{table}

Next, we evaluate our classification models and \mname technique. For each experiment, we have trained our models for 100 epochs, saving the model with highest accuracy on the testing set.

    \subsubsection{Classification}

To get a sense of the difficulty of the task, the performance of our models and various baselines has been evaluated in the following two scenarios:
\begin{itemize}
  \item{Known patients:} with $D_{20k}^{known}$, where patients in the training are also present in the testing set, we first assess the ability of the model to classify known patients, which can be useful in some clinical situations such as long-term monitoring at ICU. In this setting, clinicians wish to capture seizure patterns similar to those observed previously, or in long-ambulatory long-term monitoring settings with implantable electrodes where there is the opportunity to fine-tune the system based on previous seizures.  
  \item{Unseen patients:} using $D_{20k}^{unseen}$, we evaluate the performance of the model at generalizing across patients. It is important to ensure the model can classify EEG from different brains. This is an obviously harder task, leading to an understandable drop in accuracy when evaluating in this setting.
\end{itemize}

For each scenario, we have experimented with two variations of \mname with each $N = 512$ reference embeddings, either with supervised embedding (\texttt{\mname-CNN}) or unsupervised embedding (\texttt{\mname-CAE}), as well as the following baseline models:
\begin{itemize}
  \item{Convolutional Neural Network (\texttt{CNN}):} this baseline is the model that we use as a supervised embedding function $f$, introduced in~\ref{sec:cnn}. Therefore, the complexity of this baseline is comparable to that of \texttt{\mname-CNN}.
  \item{MultiLayer Perceptron (\texttt{MLP})}: we have trained an MLP with 1024 neurons in the hidden layer to perform the same classification task.
\end{itemize}

The results of this experiment, shown in table~\ref{tab:accuracy}, confirm the existence of inconsistencies within the dataset. When labeling the sequences, experts usually remain consistent for sequences coming from the same patient. However, it is hard for them to remain consistent when labeling sequences from new patients having their own specific patterns. This is where re-labeling with machine feedback will show how \mname can be used to drastically improve performance.

\begin{table}
\center
  \caption{Accuracy obtained on the testing set of $D_{20k}^{unseen}$. Accuracies are shown before and after re-evaluation. During co-learning, 837 sequences (4.18\% of the dataset $D_{20k}^{unseen}$) have been re-evaluated by an expert. Results are shown for the full testing set, as well as on the subset of sequences re-evaluated during co-learning. We note a clear increase in model performance after this first iteration alone.}
  \vspace{0.5cm}
  \label{tab:machfeed}
  \begin{tabular}{ccccc}
    \toprule
    \multirow{2}{*}{Model}&\multicolumn{2}{c}{Before re-labeling}&\multicolumn{2}{c}{After re-labeling}\\
    &full test&re-eval only&full test&re-eval only\\
    \midrule
    \mname-\texttt{CNN}&39.36\%&7.03\%&40.75\%&68.75\%\\
    \mname-\texttt{CAE}&38.46\%&10.94\%&39.06\%&67.97\%\\
    \texttt{CNN}&38.89\%&6.25\%&41.58\%&68.75\%\\
    \texttt{MLP}&21.04\%&0.78\%&23.14\%&14.06\%\\
    \bottomrule
  \end{tabular}
\end{table}

    \subsubsection{Co-Learning}

Results on the original $D_{20k}^{unseen}$ dataset (before label re-evaluation) from the previous experiment are again shown on the first two columns of table~\ref{tab:machfeed}. These accuracies set the starting point for improvements using co-learning.

In order to evaluate our co-learning framework, we ran one iteration of our algorithm, providing a suggestion of sample points to be re-labeled by a human expert, using the highest confidence heuristic introduced in section~\ref{sec:feedbstrategies}. Results show that after this first iteration alone, during which 837 sequences (4.18\% of the dataset $D_{20k}^{unseen}$) have been re-evaluated by an expert, accuracy on the testing shows great improvement. The results are shown on table~\ref{tab:machfeed}. We note that in most scenarios, \texttt{\mname-CNN} performs better than the unsupervised embedding alternative \texttt{\mname-CAE}, and also better than \texttt{MLP} which cannot learn properly, as expected. Our \texttt{CNN} has good performance overall, but cannot claim to be interpretable like \texttt{\mname-CNN}.

\begin{figure}
\center
\includegraphics[width=2.7in]{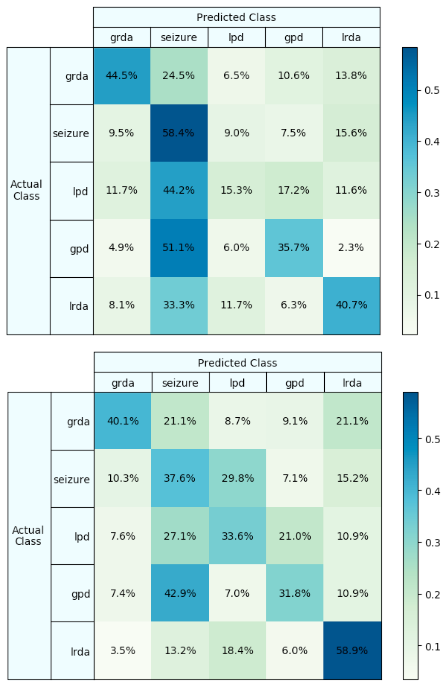}
\caption{Confusion matrices of \texttt{HAMLET-CNN} trained and tested on the original testing set (top) and after fine-tuning, with training and testing on the improved datasets (bottom). Each row shows, for a given expert-provided class label, how it has been classified and in what proportions by the model (a diagonal matrix is ideal). After co-training, although seizures are not as clearly classified as before, LRDA and LPD are not as confused with seizures anymore.}
\label{fig:confmats}
\end{figure}

The confusion matrices for \texttt{\mname-CNN} in the first iteration and after re-labeling are shown together on figure~\ref{fig:confmats}. It is interesting to notice how the model improves based on these confusion matrices: although after co-learning, seizures are not as clearly recognized by the model as before, the model now better classifies LRDA and LPD, leading to overall better accuracy and classification performance of the model.

    \subsubsection{Interpretability}

Finally, we analyzed how the model uses the reference embeddings when making decisions. This can be done by looking at the weights of the dense layer that are applied to the reference embeddings. For each output class label $c$, we selected the 16 reference embeddings with biggest weights in the dense layer, among the 512 available embeddings. We then computed what percentage of these 16 embeddings have the same class label $c_i$ as $c$. Each percentage gives an interpretability score for label $c$.

The results presented in table~\ref{table:interp} for \texttt{\mname-CNN} show that for most classes, the model is really able to learn how to use the similarity scores. Interestingly, the less interpretable classes in the model also are the ones with worst classification performance according to the confusion matrices. One reason could potentially be that the set of patterns that represent such classes is really diverse, preventing the model from efficiently using similarity scores.

Knowing that the model knows how to make use of reference embeddings, the expert re-evaluation task can benefit from interpretability outputs from the model. For a given input sequence that our model suggests as needing re-evaluation, the embeddings with highest similarity scores are selected and displayed next to the sequence, explaining why the model made that particular decision.

\begin{table}
  \center
  \caption{Interpretability scores for \texttt{HAMLET-CNN}. Each percentage shows, out of the 16 reference embeddings with highest weights in the model, how many belong to the target class.}
    \vspace{0.5cm}
  \label{table:interp}
  \begin{tabular}{lccccc}
    \toprule
    Class&LRDA&GRDA&LPD&Seizure&GPD\\
    \midrule
    Interpretability&75.00\%&68.75\%&62.50\%&37.50\%&25.00\%\\
    \bottomrule
  \end{tabular}
\end{table}

    \subsubsection{Comparison of Machine Feedback Strategies}
    
Finally, we compared the three different strategies introduced in~\ref{sec:feedbstrategies}. For each strategy, \mname suggested 128 samples with new labels. We counted how many the expert agrees with in each case. The results in table~\ref{table:feedstrats} show that entropy might be a better indicator of misclassified inputs, although all methods perform almost equally well.

\begin{table}
  \center
  \caption{Percentage of expert agreement on suggestions from \texttt{HAMLET-CNN} for various machine feedback strategies.}
    \vspace{0.5cm}
  \label{table:feedstrats}
  \begin{tabular}{lccccc}
    \toprule
    Method&Agreement\\
    \midrule
    High Confidence&92.19\%\\
    Margin Sampling&91.41\%\\
    Entropy&93.75\%\\
    \bottomrule
  \end{tabular}
\end{table}


\section{Conclusion}

By acknowledging the difficulty of label acquisition in new domains for both human experts and machine algorithms, we have reached with \mname multiple advances in deep learning for healthcare applications. To summarize, first, we have introduced a novel technique, \mname, for human and machine co-learning that is suited for creating high-quality labeled datasets on challenging tasks with a limited budget. This technique has benefits that can appreciated in many deep learning applications. We have shown how this technique applies to the classification of seizures from continuous EEG, a challenging task for both machines and human experts. Finally, we have designed a new kind of network with increased interpretability potential. During the dataset improvement phase, this interpretability via similar reference examples can assist experts re-evaluating sample sequences by explaining the reasons why the algorithm came up with a given output.

Ultimately, this work can benefit others domains that share similar difficulties when it comes to efficiently labeling large volumes of data for challenging applications of deep learning.

\bibliographystyle{ACM-Reference-Format}
\bibliography{biblio}

\end{document}